\documentclass[10pt,twocolumn,letterpaper]{article}

\usepackage{times}
\usepackage{epsfig}
\usepackage{graphicx}
\usepackage{amsmath}
\usepackage{amssymb}
\usepackage{bm}
\usepackage{geometry}

\usepackage{amsfonts}
\usepackage{amsmath}
\usepackage{amsthm}
\usepackage{widetext}
\usepackage[OT1]{fontenc} 
\usepackage{subfig}
\usepackage{graphicx}
\usepackage{xcolor}
\usepackage{placeins}
\usepackage{tabularx}
\usepackage{amsopn}
\usepackage{algpseudocode}

\newtheorem{prop}{Proposition}
\newtheorem{defn}{Definition}

\newtheorem{theorem}{Theorem}
\newcommand{\bbR}{\mathbb{R}}
\newcommand{\ones}{\mathbf{1}}
\newcommand{\OT}{\mathrm{OT}}
\newcommand{\KL}{\mathrm{KL}}
\newcommand{\Ent}{\mathrm{H}}
\newcommand{\inner}[1]{{\left\langle #1 \right\rangle}}

\renewcommand{\vec}[1]{\bm{#1}}

\usepackage[pagebackref=true,breaklinks=true,letterpaper=true,colorlinks,bookmarks=false,hypertexnames=false]{hyperref}

\begin{document}

\title{A unified framework for non-negative matrix and tensor factorisations with a smoothed Wasserstein loss}

\author{Stephen Y. Zhang\\
Department of Mathematics, University of British Columbia\\
Vancouver, BC Canada\\
{\tt\small syz@math.ubc.ca}
}
\date{}

\maketitle

\begin{abstract}
    Non-negative matrix and tensor factorisations are a classical tool for finding low-dimensional representations of high-dimensional datasets. In applications such as imaging, datasets can be regarded as distributions supported on a space with metric structure. In such a setting, a loss function based on the Wasserstein distance of optimal transportation theory is a natural choice since it incorporates the underlying geometry of the data. We introduce a general mathematical framework for computing non-negative factorisations of both matrices and tensors with respect to an optimal transport loss. We derive an efficient computational method for its solution using a convex dual formulation, and demonstrate the applicability of this approach with several numerical illustrations with both matrix and tensor-valued data.
\end{abstract}

\section{Introduction}

Matrix and tensor factorisations are a classical tool for extracting low-dimensional structure from complex high-dimensional datasets. The seminal work of Lee and Seung \cite{lee1999learning} noted that real-world datasets are often naturally non-negative, and introduced non-negative matrix factorisation (NMF) for finding an approximate low-rank representation of a matrix-valued dataset that is easier to interpret. NMF can be interpreted in a sparse-coding sense, in that a small set of atoms is sought that can approximately generate the full dataset in its non-negative linear span. This general concept of finding linear representations in terms of a small number of components also extends to tensors \cite{welling2001positive}, for which many notions of non-negative decompositions have been proposed, including the popular CANDECOMP/PARAFAC (CP) format \cite{kolda2009tensor}.

Approximate factorisations are typically sought with respect to some divergence function \cite{kolda2009tensor, kim2007nonnegative} such as the squared Frobenius norm or Kullback-Leibler divergence. Such divergences decompose elementwise in their matrix or tensor-valued arguments. In settings such as imaging the observed data lie naturally on a metric space, and elementwise divergences cannot take advantage of this additional structure. Recent works \cite{rolet2016fast, qian2016non, schmitz2018wasserstein, sandler2011nonnegative} have focused on addressing this issue in the context of NMF by employing a Wasserstein loss that accounts for the geometry of the data by using optimal transport.  

In this work, we generalise the smoothed dual approach of Rolet et al. \cite{rolet2016fast} from matrices to the setting of tensors. The problem of finding non-negative tensor factorisations with a Wasserstein loss has remained untouched until very recently, when it was addressed by the work of Afshar et al. \cite{afshar2020swift} in which a non-negative CP decomposition was sought via the primal formulation of optimal transport \cite{frogner2015learning}. In contrast, the approach we consider proceeds via convex duality in order to take advantage of the availability of closed-form gradients in the dual problem  \cite{cuturi2016smoothed, cuturi2018semidual, rolet2016fast}. 

Our work presents a unified framework for Wasserstein factorisation problems, since it handles the fully general case of finding a Tucker decomposition and includes non-negative CP decompositions and NMF as special cases.

\section{Background}\label{sec:background}
The reader is provided with an overview of our notation conventions in Supplement \ref{sec:notation}.

\subsection{Non-negative matrix factorisation}
As a prelude to the more general problem of non-negative tensor factorisations, we discuss the case of NMF. Given a $m \times n$ non-negative matrix $X \in \bbR^{m \times n}_{\ge 0}$ and a target rank $1 \leq r \leq \min(m, n)$, the NMF problem \cite{lee1999learning, wang2012nonnegative} seeks to find non-negative factor matrices $U \in \bbR^{m \times r}_{\ge 0}, V \in \bbR^{n \times r}_{\ge 0}$ such that we have a rank-$r$ approximation $X \approx UV^\top = \sum_{k = 1}^r U_k \otimes V_k$
in some appropriate sense. In terms of the columns of $X$, we have equivalently
\begin{align}
    X_i \approx (UV^\top)_i = \sum_{k = 1}^r U_k V_{ik}. \label{eq:nmf_basis_rep}
\end{align}
If columns of $X$ are observations, then \eqref{eq:nmf_basis_rep} represents each observation $X_i$ as a linear combination of $r$ atoms $\{ U_k \}_{k = 1}^r$ with non-negative coefficients $\{ V_{ik} \}_{k = 1}^r$. The factor matrix $U$ thus contains an approximate $r$-element basis for the dataset.

Factors $U$ and $V$ are found by solving a minimisation problem of the form
\begin{align}
    \min_{U \in \bbR^{m \times r}_{\ge 0}, V \in \bbR^{n \times r}_{\ge 0}} \varphi(X, UV^\top).
\end{align}
In the above, $\varphi(\cdot, \cdot)$ is a suitably chosen loss function over matrices, commonly taken to be the previously mentioned squared Frobenius norm $\varphi(A, B) = \| A - B \|_F^2$ or the generalised Kullback-Leibler (KL) divergence $\varphi(A, B) = \KL(A | B)$ \cite{wang2012nonnegative}.

\subsection{Non-negative tensor factorisation}\label{sec:ntf}

\begin{figure}[h]
    \centering
    \includegraphics[width = 0.75\linewidth]{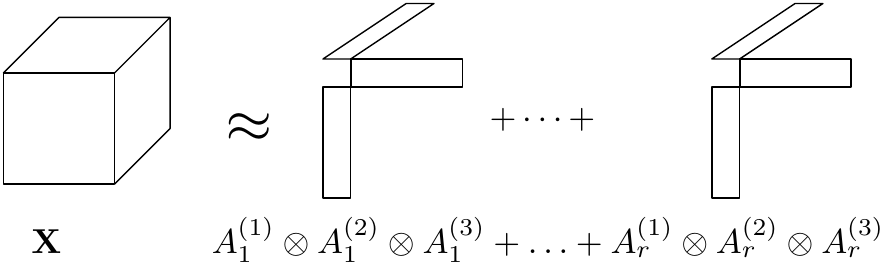}
    \caption{Illustration of the CP tensor decomposition format.}
    \label{fig:cp_concept}
\end{figure}

Now consider a non-negative $d$-mode tensor $X \in \bbR^{n_1 \times \cdots \times n_d}_{\ge 0}$, for which we seek a low-rank, non-negative representation in a similar sense to NMF. The NMF problem can be directly generalised to $d$-mode tensors to yield the CP decomposition format \cite{kolda2009tensor}. Given a target rank $r$, we seek $d$ factor matrices $A^{(i)} \in \bbR^{n_i \times r}_{\ge 0}, 1 \leq i \leq d$ such that 
\begin{align}
    X \approx [A^{(1)}, \ldots, A^{(d)}] := \sum_{i = 1}^r A^{(1)}_i \otimes \cdots \otimes A^{(d)}_i.
\end{align}
That is, $X$ can be approximated as a sum of $r$ rank-1 tensors, which we illustrate in Figure \ref{fig:cp_concept}. For $d = 2$, one recovers the rank-$r$ NMF problem. 

The Tucker decomposition format further generalises the CP format \cite{kolda2009tensor}. Given a tensor $X$ and a $d$-tuple $(r_1, \ldots, r_d)$ specifying the \emph{multilinear rank} of the decomposition, one seeks a core (also known as \emph{mixing}) tensor $S \in \mathbb{R}^{r_1 \times \cdots \times r_d}_{\ge 0}$ and factor matrices $A^{(i)} \in \bbR^{n_i \times r_i}_{\ge 0}$, $1 \leq i \leq d$ such that 
\begin{align}
    \begin{split}
        X &\approx S[A^{(1)}, \cdots, A^{(d)}] \\
        &:= \sum_{i_1 = 1}^{r_1} \cdots \sum_{i_d = 1}^{r_d} S_{i_1, \ldots, i_d} A^{(1)}_{i_1} \otimes \cdots \otimes A^{(d)}_{i_d}.
    \end{split}
\end{align}
This can be interpreted similarly to the CP format, but with the core tensor $S$ encoding interactions between the columns of the factor matrices $A^{(i)}$. Importantly, when $r_1 = \cdots = r_d = r$ and $S = \delta_{i_1, \ldots, i_d}$, we recover the CP format. 

\subsection{Optimal transport}\label{sec:ot}

Optimal transport (OT) deals with comparison of (probability) distributions supported on spaces with metric structure. Wasserstein distances have has found broad applications in statistics and machine learning in recent years \cite{peyre2019computational}, since they are sensitive to ``horizontal'' displacements, in contrast to other commonly used distances or divergences that are typically only sensitive to ``vertical'' discrepancies of distributions. 

The central optimal transport problem is: given $\alpha, \beta$ probability distributions and a matrix $C$ measuring the cost $C_{ij}$ of transport from point $i$ to point $j$, find the \emph{coupling} $\gamma$ (i.e. joint distribution having marginals $\alpha$ and $\beta$) solving 
\begin{align}
    \OT(\alpha, \beta) := \inf_{\gamma \in \Gamma(\alpha, \beta)} \inner{C, \gamma}. \label{eq:monge_kantorovich}
\end{align}
where $\Gamma(\alpha, \beta) = \{\gamma \ge 0 : \gamma \ones = \alpha, \gamma^\top \ones = \beta \}$ denotes the set of all couplings of $(\alpha, \beta)$. 
Importantly, in the case where $\alpha, \beta$ are supported on a space $\mathcal{X}$ with a distance function $d_\mathcal{X}$, one may pick $C(x, y) = d_\mathcal{X}(x, y)^2$. Then $(\alpha, \beta) \mapsto \OT(\alpha, \beta)$ establishes a natural metric on the space of probability distributions with finite second moment known as the 2-Wasserstein ($W_2$) metric (and more generally, one can define the $p$-Wasserstein metric) \cite[Proposition 2.2]{peyre2019computational}. 

In practice, the entropic regularisation \cite{cuturi2013sinkhorn}, \cite[Chapter 4]{peyre2019computational} is often employed instead:
\begin{align}\label{eq:ent_ot}
    \begin{split}
        \OT_\varepsilon(\alpha, \beta) :=& \inf_{\gamma \in \Gamma(\alpha, \beta)} \inner{C, \gamma} + \varepsilon E(\gamma) \\ 
        =& \inf_{\gamma \in \Gamma(\alpha, \beta)} \varepsilon \Ent(\gamma | e^{-C/\varepsilon}) ,
    \end{split}
\end{align}
where $\varepsilon > 0$ is the regularisation parameter. In the limit $\varepsilon \to 0^+$, the solution of \eqref{eq:ent_ot} converges to that of \eqref{eq:monge_kantorovich} \cite[Proposition 4.1]{peyre2019computational}. The problem \eqref{eq:ent_ot} can be solved efficiently using methods such as the Sinkhorn algorithm \cite{cuturi2013sinkhorn}. Since it is smooth and strictly convex, it is commonly used as a loss function that approximates the Wasserstein distance \cite{cuturi2016smoothed, rolet2016fast, schmitz2018wasserstein}. 

A shortcoming of the typical formulations of optimal transport (\ref{eq:monge_kantorovich}, \ref{eq:ent_ot}) is that the problem is posed over normalised distributions. In practice, input data may not be perfectly normalised, and rescaling could be undesirable, resulting in a potential loss of information. Various works \cite{chizat2018scaling, frogner2015learning, liero2018optimal} consider relaxations of optimal transport to deal with the case where $\alpha, \beta$ are allowed to be positive measures. We introduce here \emph{semi-unbalanced} transport, which takes the form
\begin{align} \label{eq:semiunbal_ot_def}
    \OT_\varepsilon^\lambda(\alpha, \beta) &= \inf_{\gamma : \gamma \ones = \alpha} \varepsilon \Ent(\gamma | e^{-C/\varepsilon}) + \lambda \KL(\gamma^\top \ones | \beta). 
\end{align}
Here, the transport plan $\gamma$ is required to agree only approximately with the second input measure via the soft marginal penalty $\KL(\cdot | \beta)$. The parameter $\lambda$ controls the strength of the soft marginal constraint: sending $\lambda \to +\infty$, we recover the standard entropy-regularised problem \eqref{eq:ent_ot}. Like its balanced counterpart, unbalanced OT problems can be solved using a generalised Sinkhorn-like scheme \cite{chizat2018scaling}.

\section{Wasserstein tensor factorisation}

\subsection{Optimal transport as a distance on tensors}\label{sec:ot_tensor_def}

Optimal transport deals with distributions supported on spaces with metric structure, as introduced in Section \ref{sec:ot}. Tensors can be naturally cast in this framework by thinking in terms of a product of metric spaces. In concrete terms, suppose the $i$th mode of the tensor $X \in \mathbb{R}^{n_1 \times \cdots \times n_d}$ corresponds a discrete metric space $(\mathcal{X}_i, d^{(i)})$. Then, the tensor $X$ lives on a product of metric spaces $(\mathcal{X}, d_\mathcal{X}) = (\mathcal{X}_1 \oplus \mathcal{X}_2 \oplus \cdots \oplus \mathcal{X}_d, d_\mathcal{X})$. 

In general, the choice of the product metric $d_\mathcal{X}$ is not unique. Let us consider, however, the $p$-product metric \cite{deza2009encyclopedia} on $\mathcal{X}$ for $1 \leq p < \infty$:
\begin{align}
    d_\mathcal{X}(\vec{x}, \vec{y})  &= (d^{(1)}(x_1, y_1)^p + \cdots + d^{(d)}(x_d, y_d)^p)^{\frac{1}{p}}.\label{eq:prod_metric}
\end{align}
This family of product metrics leads to a convenient formulation of optimal transport: to formulate the $p$-Wasserstein distance on a product of discrete spaces, we need to form the cost tensor $C$ encoding cost for moving an unit mass from $(i_1, \ldots, i_d)$ to $(j_1, \ldots, j_d)$. For $X$ a tensor with $d$ modes, $C$ has $2d$ modes and decomposes additively:
\begin{align}
    \begin{split}
        C_{i_1, \ldots, i_d, j_1, \ldots, j_d} &= d_\mathcal{X}((i_1, \ldots, i_d), (j_1, \ldots, j_d))^p \\
        &= d^{(1)}(i_1, j_1)^p + \cdots + d^{(d)}(i_d, j_d)^p \\
        &= C^{(1)}_{i_1, j_1} + \cdots + C^{(d)}_{i_d, j_d},
    \end{split}\label{eq:cost_tensor}
\end{align}
where $C^{(k)}$ is the cost matrix corresponding to the $k$th mode of the tensor $X$. We thus define the Wasserstein distance between tensors $X, Y \in \mathcal{P}(\mathcal{X})$ to be 
\begin{align}
    \OT(X, Y) &:= \inf_{\gamma \in \Gamma(X, Y)} \inner{C, \gamma},\label{eq:ot_mk_tensor}
\end{align}
where $\Gamma(X, Y)$ denotes the set of all possible couplings of the tensors $X, Y$, i.e. \begin{align}
    \begin{split}
        \Gamma(X, Y) = \{ &\gamma \in \mathcal{P}(\mathcal{X} \otimes \mathcal{X}) :  \\
        &\sum_{j_1, \ldots, j_d} \gamma_{i_1, \ldots, i_d, j_1, \ldots, j_d} = X_{i_1, \ldots, i_d},  \\
        & \sum_{i_1, \ldots, i_d} \gamma_{i_1, \ldots, i_d, j_1, \ldots, j_d} = Y_{j_1, \ldots, j_d} \}.
    \end{split}
\end{align}

\subsection{Problem setup}

We will first discuss the problem of finding non-negative tensor factorisations in the fully general case of a Tucker decomposition, since the settings of CP tensor decompositions and NMF fit naturally in this framework as special cases (see Supplement \ref{sec:wnmf}). Our approach to the Wasserstein tensor factorisation (WTF) problem is based on the approach to NMF introduced by Rolet et al. \cite{rolet2016fast}, where the authors utilise duality properties of entropically smoothed optimal transport to efficiently solve a series of smooth, convex subproblems for the factor matrices. 

As previously, we consider a non-negative $d$-mode tensor $X \in \bbR^{n_1 \times \cdots \times n_d}_{\ge 0}$, for which we seek a Tucker decomposition $S[A^{(1)}, \ldots, A^{(d)}]$, where $A^{(i)} \in \bbR^{n_i \times r_i}_{\ge 0}$ is the $i$th factor matrix and $S \in \bbR^{r_1 \times \cdots \times r_d}_{\ge 0}$ is the core tensor. Let $\Phi : \bbR^{n_1 \times \cdots \times n_d}_{\ge 0} \times \bbR^{n_1 \times \cdots \times n_d}_{\ge 0} \to [0, \infty)$ be a loss function on tensors based on optimal transport, which we define in further detail in Section \ref{sec:ot_loss_func}. For now, we will require that $\Phi$ is smooth and convex in its second argument.

The fundamental WTF problem can be written then as 
\begin{align}
    \min_{S, A^{(1)}, \ldots, A^{(d)} \ge 0} \Phi(X, S[A^{(1)}, \ldots, A^{(d)}]).
\end{align}
In addition, we may optionally impose normalisation constraints on the decomposition components. This is a natural constraint for such settings where $X$ contains histogram or count data, since it resolves the issue of multiplicative non-uniqueness in the scaling of factor matrices (that is, without normalisation constraints the decomposition does not change when any two factor matrices are multiplied by $\kappa$ and $\kappa^{-1}$ respectively). We list some useful normalisation constraints in Table \ref{table:norm}. For ease of notation, we will denote by $\Sigma_i$ the constraint set for the $i$th factor matrix, and $\Sigma_0$ that of the core tensor.
\begin{table}
    \begin{tabular}{|l|l|}
        \hline
        Choice of normalisation & $\Sigma$ \\ 
        \hline
        Row-normalised factor matrices &  $\{ A^{(i)} \ones = \ones \}$ \\ 
        Column-normalised factor matrices & $ \{ (A^{(i)})^\top \ones = \ones \}$ \\ 
        Fully normalised factor matrices & $ \{ \inner{A^{(i)}, \ones} = 1 \} $ \\ 
        Fully normalised core tensor & $\{ \inner{S, 1} = 1 \}$  \\ \hline
    \end{tabular}
    \caption{Normalisation constraints on factors}
    \label{table:norm}
\end{table}

Following \cite{rolet2016fast}, we relax the non-negativity constraint by using a entropy barrier function. For future convenience (see Sections \ref{sec:factors} and \ref{sec:core}), we choose to incorporate normalisation constraints into the barrier function: we define $E_{\Sigma_i}(x) = E(x) + \iota(x \in \Sigma_i)$, where $\iota$ denotes the indicator function of a convex set 
\begin{align}
    \iota(x \in A) &= \begin{cases}
            0, &x \in A; \\ 
            +\infty, &\text{otherwise}.
    \end{cases}.
\end{align}
This yields the following smooth and unconstrained WTF problem
\begin{align}
    \begin{split}
        \min_{S, A^{(1)}, \ldots, A^{(d)}} &\Phi(X, S[A^{(1)}, \ldots, A^{(d)}]) \\ 
        &+ \rho_0 E_{\Sigma_0}(S) + \sum_{i = 1}^d \rho_i E_{\Sigma_i}(A^{(i)}), \label{eq:wtf_general_smooth}
    \end{split}
\end{align}
and this is the form of the problem which we seek to solve. 
The problem \eqref{eq:wtf_general_smooth} is convex individually in each of the factors, but not jointly. We may therefore seek a local minimum by performing a block coordinate descent in each of the core tensor and factor matrices. For each convex subproblem, we proceed via convex duality \cite{rockafellar1967}. We state now a result on the Legendre transform (see Supplement \ref{sec:convex_duality} for a definition) of the entropy functional \cite{rolet2016fast, cuturi2016smoothed}.

\begin{prop}[Legendre transform of entropy]\label{prop:legendre_entropy}
    The Legendre transform of the unconstrained entropy $\alpha \mapsto E(\alpha)$ is $E^*(u) = \inner{\exp(u), \ones}$. 
    
    Now let $\Sigma$ be a constraint set. Then up to a constant, the Legendre transform of the constrained entropy $\alpha \mapsto E_\Sigma(\alpha)$ is:
    \begin{align}
        E_\Sigma^*(u) &= \begin{cases}
            \log\inner{\exp(u), \ones}, & \Sigma = \{ \alpha : \inner{\alpha, \ones} = 1 \} \\
            \sum_i \log \inner{\exp(u_i), \ones}, & \Sigma = \{ \alpha : \alpha^\top \ones = \ones \} \\
            \sum_i \log \inner{\exp((u^\top)_i), \ones}, &\Sigma = \{ \alpha : \alpha \ones = \ones \}
        \end{cases}
    \end{align}
    We note that the first definition applies for vector, matrix or tensor-valued $\alpha$, whilst the latter two apply only to matrix-valued $\alpha$. 
\end{prop}

We now formulate the dual problems for the factor matrices and core tensor. Proofs are deferred to Supplement \ref{sec:proofs}. 

\subsection{Tensor decompositions -- factor matrices}\label{sec:factors}

For the moment let $S$ be held fixed, and we seek to optimise over only the factor matrices $A^{(k)}$. We recall from Section \ref{sec:ntf} that in the case where $r_1 = \cdots = r_d = r$ and $S_{i_1, \ldots, i_d} = \delta_{i_1, \ldots, i_d}$, this corresponds exactly to the CP decomposition format \cite{kolda2009tensor}. 

We consider the convex and formally unconstrained subproblem for a single factor matrix $A^{(k)}$,
\begin{align} \label{eq:primal_factors}
    \begin{split}
        &\min_{A^{(k)}} \Phi(X, S[A^{(1)}, \ldots, A^{(d)}]) + \rho_k E_{\Sigma_k}(A^{(k)}).  \\
    \end{split}
\end{align}
The corresponding dual problem is given by the following Proposition.

\begin{prop}[Dual problem for factor matrices]\label{prop:factor_duality}~\\
The dual problem for the $k$th factor matrix $A^{(k)}$, corresponding to \eqref{eq:primal_factors}, is 
\begin{align}\label{eq:dual_factors}
    \min_{U \in \mathbb{R}^{n_1 \times \cdots \times n_d}} \Phi^*(X, U) + \rho_k E_{\Sigma_k}^*\left( \frac{-1}{\rho_k} \Xi^{(k)}(U) \right),
\end{align}
where we have written $\Xi^{(k)}$ to be a linear function of $U$:
\begin{align}
    \Xi^{(k)}(U) &= \left[ U \times_{j \ge k+1} (A^{(j)})^\top \right]_{(k)} \left[ S \times_{j \leq k-1} A^{(j)} \right]^\top_{(k)}.
\end{align}
This problem is smooth and convex in the variable $U$. 

Furthermore, at optimality, the primal variable ${A^{(k)}}^\star$ can be recovered from the optimal dual variable $U^\star$ as the solution of 
\begin{align}
    \sup_{A^{(k)}} \frac{-1}{\rho_k} \inner{A^{(k)}, \Xi^{(k)}(U^\star)} - E_{\Sigma_k}(A^{(k)}).
\end{align}
In particular, letting $Z = \exp\left( \frac{-1}{\rho_k} \Xi^{(k)}(U^\star) \right)$, we get 
\begin{align}
    {A^{(k)}}^\star &= \begin{cases}
                    Z, &\Sigma_k = \{ \}, \\
                    Z/\inner{Z, \ones}, &\Sigma_k = \{ A^{(k)} : \inner{A^{(k)}, \ones} = 1 \}, \\ 
                    \mathrm{diag}(Z \ones)^{-1} Z , &\Sigma_k = \{ A^{(k)} : A^{(k)} \ones = 1 \} \\ %
                    Z \mathrm{diag}(Z^\top \ones)^{-1} , &\Sigma_k = \{ A^{(k)} : (A^{(k)})^\top \ones = 1 \} %
                \end{cases}
\end{align}
\end{prop}
For the choices of $\Phi$ and $\Sigma_k$ that we consider, the dual problem \eqref{eq:dual_factors} is a smooth, unconstrained and convex problem in the dual variable $U$, and can thus be solved using general gradient-based methods. 

\subsection{Tensor decompositions -- core tensor}\label{sec:core}

Now we will consider optimising over the core tensor $S$, for fixed factor matrices $A^{(i)}$. The subproblem for $S$ reads
\begin{align}\label{eq:primal_core}
    \begin{split}
        &\min_{S} \Phi(X, S[A^{(1)}, \ldots, A^{(d)}]) + \rho_0 E_{\Sigma_0}(S)  \\
    \end{split}
\end{align}
We state now the corresponding dual problem.

\begin{prop}\label{prop:core_duality}
    The dual problem corresponding to \eqref{eq:primal_core} is 
    \begin{align}
        \min_{U \in \mathbb{R}^{n_1 \times \cdots \times n_d} } \Phi^*(X, U) + \rho_0 E_{\Sigma_0}^*\left( \frac{-1}{\rho_0} \Omega(U) \right),
    \end{align}
    where we have written $\Omega(U) = U \times_{j = 1}^d (A^{(j)})^\top$.
    This problem is smooth and convex in the variable $U$. 
    
    Furthermore, at optimality, the primal variable $S$ can be recovered from the optimal dual variable $U^\star$ as the solution of 
    \begin{align}
        \sup_S \inner{ \frac{-1}{\rho_0} \Omega(U^\star), S} - E_{\Sigma_0}(S).
    \end{align}
    In particular, letting $Z = \exp\left( \frac{-1}{\rho_0} \Omega(U^\star) \right)$, we get 
    \begin{align}
        S^\star &= \begin{cases}
            Z, &\Sigma_0 = \{ \} \\
            Z/\inner{Z, \ones}, &\Sigma_0 = \{ S : \inner{S, \ones} = 1 \}
        \end{cases}.
    \end{align}
\end{prop}
\noindent As with Section \ref{sec:factors}, this is a smooth, unconstrained and convex problem in the dual variable $U$. 

\subsection{Legendre transform for optimal transport loss}

The strategy we proposed in Sections \ref{sec:factors}, \ref{sec:core} relies on having access to the Legendre transform of $\Phi(X, \cdot)$. We state now two crucial results about the Legendre transform of $\beta \mapsto \OT_\varepsilon(\alpha, \beta)$ and $\beta \mapsto \OT^\lambda_\varepsilon(\alpha, \beta)$, which will allow us to compute $\Phi^*(X, \cdot)$ for certain choices of $\Phi$. 

\begin{prop}[Semi-unbalanced transport]\label{prop:semiunbal_dual}
    The Legendre transform of $\beta \mapsto \OT^\lambda_\varepsilon(\alpha, \beta)$ is 
    \begin{align}\label{eq:semiunbal_ot_legendre}
        {\OT_\varepsilon^\lambda}^* (\alpha, u) &= -\varepsilon \inner{\alpha, \log(\alpha \oslash (K f(u))) - \ones}, 
    \end{align}
    where $f(u) = \left( \frac{\lambda}{\lambda - u} \right)^{\lambda/\varepsilon}$,
    and at optimality, the relationship between the primal variable $\beta^\star$ and dual variable $u^\star$ is 
    \begin{align}
        \beta^\star &= \left( \frac{\lambda}{\lambda - u^\star} \right) f(u^\star) \odot K^\top \frac{\alpha}{Kf(u^\star)}.
    \end{align}
    where in the above we have written $K = e^{-C/\varepsilon}$ to be the Gibbs kernel \cite{peyre2019computational}. 
\end{prop}
\begin{proof}
    See Supplement \ref{sec:proofs}.
\end{proof}

Since we recover $\OT_\varepsilon$ from $\OT^{\lambda}_\varepsilon$ in the limit $\lambda \to +\infty$, we expect their Legendre transforms to also coincide in the limit. We note that for Proposition \ref{prop:semiunbal_dual}, in the limit $\lambda \to +\infty$ we have:
\begin{align}
    \lim_{\lambda \to +\infty} \left( \frac{\lambda}{\lambda - u} \right)^{\lambda/\varepsilon} = e^{u/\varepsilon}.
\end{align}
Thus we recover the known result for balanced transport (see e.g. \cite[Theorem 2.4]{cuturi2016smoothed}):
\begin{prop}[Balanced transport]
    The Legendre transform of $\beta \mapsto \OT_\varepsilon(\alpha, \beta)$ is 
    \begin{align}
        \OT^*_{\varepsilon}(\alpha, u) &= -\varepsilon \inner{\alpha, \log( \alpha \oslash (K e^{u/\varepsilon})) - \ones } 
    \end{align}
    Furthermore, at optimality, the relationship between the primal variable $\beta^\star$ and dual variable $u^\star$ is 
    \begin{align}
        \beta^\star &= e^{u^\star/\varepsilon} \odot K^\top \frac{\alpha}{K e^{u^\star/\varepsilon}}.
    \end{align}
\end{prop}

\subsection{Optimal transport as a loss functional}\label{sec:ot_loss_func}

We discuss now choices of the loss functional $\Phi$ which may be useful in certain contexts. Although Rolet et al. \cite{rolet2016fast} employ the well-known entropy-regularised optimal transport loss \eqref{eq:ent_ot}, we will use instead the more flexible semi-unbalanced loss \eqref{eq:semiunbal_ot_def}. 

In particular, in applications sometimes input data may not be perfectly normalised \cite{qian2016non}, or may have a multimodal distribution \cite{schmitz2018wasserstein}. In such settings, a strict requirement for mass transport may result in excessive sensitivity to noise in the input data. Further, for factorisation problems we found empirically that using the semi-unbalanced loss can improve numerical stability. This can be explained by noting that a candidate low-rank approximation may not always perfectly match the input in terms of total mass.

In the following, we work with the semi-unbalanced loss $\OT^{\lambda}_\varepsilon(\cdot, \cdot)$, since we formally recover balanced transport in the limit $\lambda \to +\infty$.

\begin{prop}[Smoothed Wasserstein loss on tensors]\label{prop:loss_tensor}
    Let $\mathcal{X}$ be a $d$-dimensional discrete product metric space as discussed in Section \ref{sec:ot_tensor_def}. Suppose $X, \hat{X} \in \mathcal{M}_+(\mathcal{X})$. Then, the smoothed Wasserstein loss directly applied on tensors is $\Phi(X, \hat{X}) = \OT^\lambda_\varepsilon(X, \hat{X})$, where $C_{i_1, \ldots, i_d, j_1, \ldots, j_d} = \sum_{k = 1}^d C^{(k)}_{i_k, j_k}$ is a cost tensor that decomposes along the modes. We write $C^{(k)}$ to be the cost matrix for the $k$th mode.
    
    Let $U \in \mathbb{R}^{n_1 \times \cdots \times n_d}$ be a tensor of dual variables corresponding to $\hat{X}$. Then we have $\Phi^*(X, U) = {\OT_\varepsilon^{\lambda}}^*(X, U)$,
    where we interpret the formula \eqref{eq:semiunbal_ot_legendre} of Proposition \ref{prop:semiunbal_dual} in terms of inner products, elementwise operations, and contractions on tensors.  
\end{prop}
\noindent A key step in evaluating ${\OT^{\lambda}_\varepsilon}^*$ is a convolution with the Gibbs kernel $K = e^{-C/\varepsilon}$. For a cost \eqref{eq:cost_tensor} that decomposes additively along different modes, this amounts to a series of $n$-mode products (see Supplement \ref{sec:conv}) for which there exist efficient parallel computation schemes \cite{kaya2016high}. Therefore, there is no need to directly deal with the (prohibitively large) full cost tensor $C$. 

For imaging applications \cite{hazan2005sparse}, 3-mode tensors arise naturally by stacking two-dimensional images. In such a setting, the obvious choice is a sum of Wasserstein losses over images, which are slices of the tensor $X$. 
\begin{prop}[Smoothed Wasserstein loss along slices]\label{prop:loss_slice}
    Let $\mathcal{X}$ be a two-dimensional discrete metric space and suppose that $X$ is a 3-mode tensor such that the slice $X_{i, \cdot, \cdot} \in \mathcal{M}_+(\mathcal{X})$ is a matrix containing 2-dimensional information such as an image. Then in terms of matricisations, the columns of $X_{(1)}^\top$ are the vectorised images and so a Wasserstein loss that decomposes along slices is 
    \begin{align}\label{eq:loss_slice}
        \Phi(X, \hat{X}) &= \sum_{i = 1}^{n_1} \OT^\lambda_\varepsilon\left(\left(X_{(1)}^\top\right)_i , \left(\hat{X}_{(1)}^\top\right)_i\right),
    \end{align}
    where $\OT^\lambda_\varepsilon$ is a smoothed Wasserstein distance between 1-dimensional histograms, with a cost matrix encoding distances between vectorised images. This approach addresses the setting of sparse image coding using tensors introduced by \cite{hazan2005sparse}.
    
    Let $U$ be a tensor of dual variables corresponding to $\hat{X}$. Note that each scalar entry of $\hat{X}$ appears only once in the sum \eqref{eq:loss_slice}. Thus, the Legendre transform decomposes along the sum and we have:
    \begin{align}
        \Phi^*(X, U) &= \sum_{i = 1}^{n_1} {\OT^\lambda_\varepsilon}^*\left(\left(X_{(1)}^\top\right)_i , \left(U_{(1)}^\top\right)_i\right).
    \end{align}
\end{prop}

\subsection{Implementation}

To find a local minimum of \eqref{eq:wtf_general_smooth}, we perform in general a block coordinate descent in $A^{(1)}, \ldots, A^{(d)}, S$. Each convex subproblem is solved via its dual problem, which admits closed form gradients \cite{cuturi2016smoothed, rolet2016fast}. We use L-BFGS for moderately sized problems, and gradient descent for large problems. We employ the PyTorch automatic differentiation engine \cite{paszke2017automatic} and the Tensorly tensor routine library \cite{tensorly}. Since the problem \eqref{eq:wtf_general_smooth} is non-convex, a good initialisation may improve the result: while a random initialisation of the core tensor and factor matrices as done by \cite{afshar2020swift} can be used, in Section \ref{sec:results} we opt to employ the commonly used non-negative SVD initialisation \cite{kolda2009tensor}. 
\section{Results}\label{sec:results}

\subsection{Simulated data -- 3-mode tensor}

We first deal with the setting where the input tensor is a histogram lying on a product of metric spaces. We take the space $\mathcal{X} = \mathcal{\widehat{X}}^{3}$ where $\widehat{\mathcal{X}}$ is a regularly spaced grid with 128 points, equipped with the squared product metric, i.e. \eqref{eq:prod_metric} with $p = 2$. The cost tensor decomposes following \eqref{eq:cost_tensor}. Cost matrices $C^{(i)}$, $i = 1, 2, 3$ are normalised to unit mean. 

\begin{figure}[h]
    \centering
    \subfloat[]{\includegraphics[width = \linewidth]{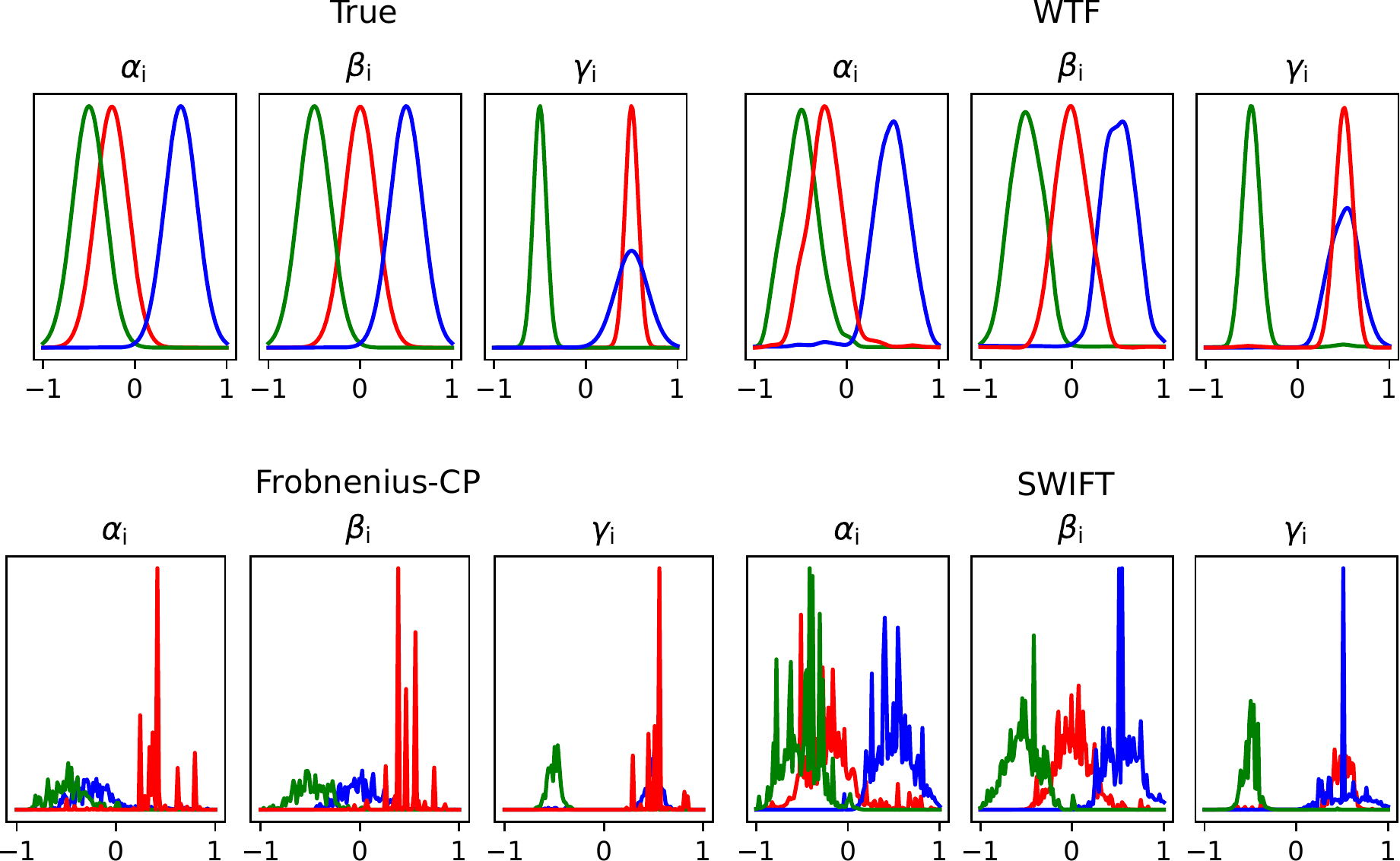}} \\ 
    \subfloat[]{\includegraphics[width = 0.98\linewidth]{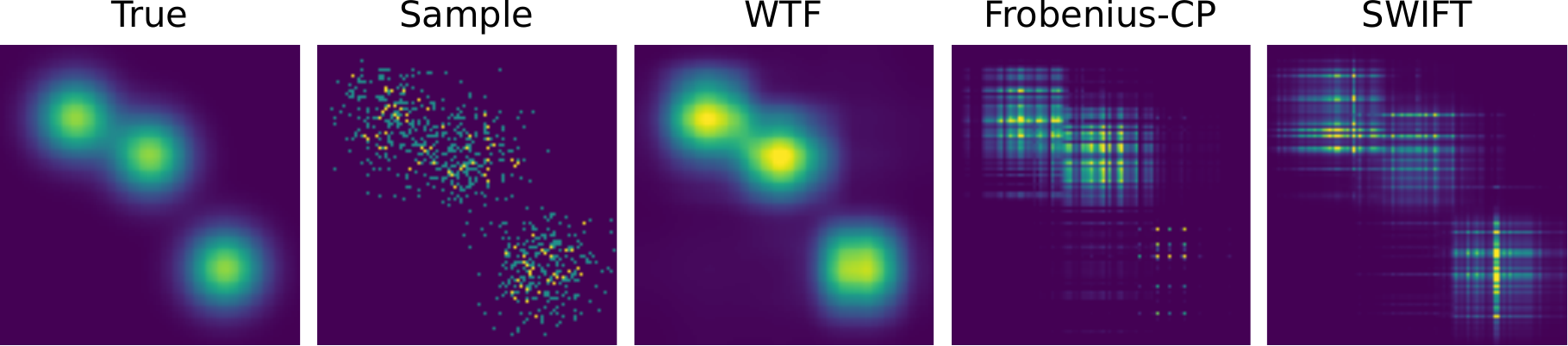}} 
    \caption{(a) True and recovered atoms visualised as univariate distributions, using WTF, Frobenius, and SWIFT decomposition methods. (b) True, sampled, and recovered tensors visualised as projections onto the first two dimensions.}
    \label{fig:3mode}
\end{figure}

We construct a ground truth tensor that is the mixture of three separable distributions: $X_\mathrm{true} = \sum_{i = 1}^3 \alpha_i \otimes \beta_i \otimes \gamma_i$, where $\{ \alpha_i, \beta_i, \gamma_i \}_{i = 1}^3$ are discrete univariate Gaussians supported on $\widehat{\mathcal{X}}$ that we illustrate in Figure \ref{fig:3mode}(a). We now consider a scenario where we have access only to limited sample observations from the ground truth. Given these samples, we seek to recover the separable components of the underlying distribution by finding a low-rank approximation to the (high-rank) observed tensor. Here, we take the tensor $X$ to be an empirical distribution drawn from $X_\mathrm{true}$. We next apply WTF with parameters $\varepsilon = 0.01, \rho_i = 10^{-3}, \lambda = 25$ to find a rank-3 approximation to $X$, with the additional constraint that the learned univariate components be normalised. For comparison, we also computed rank-3 approximations using a standard non-negative CP factorisation with a Frobenius loss, and the SWIFT algorithm \cite{afshar2020swift} with the identical parameters $\varepsilon = 0.01, \lambda = 25$. 

We show the recovered univariate factors in Figure \ref{fig:3mode}(a), and also visualise projections of the recovered tensors in Figure \ref{fig:3mode}(b). From this, we see that WTF recovers smooth atoms that are faithful to the ground truth. In contrast, Frobenius-CP finds irregular atoms that contain many `spikes' and fail to capture the underlying structure in the empirical distribution. This behaviour can be partially explained by the pointwise nature of the Frobenius norm, which is insensitive to the spatial structure in the noisy input tensor. Finally, SWIFT produces atoms which correspond roughly to the true factors (i.e. localised in the correct region). However, these are significantly more noisy than those recovered by WTF. We hypothesise that the difference in behaviour observed between WTF and SWIFT partly due to the different formulation of optimal transport on tensors -- WTF employs optimal transport with the natural product metric \eqref{eq:cost_tensor} which allows mass to be transported `globally' on the product space. In contrast, SWIFT uses a sum of optimal transport terms on the $i$-mode fibers of the tensor \cite{afshar2020swift} and thus for each term, there is the limitation that mass is constrained to be transported along one-dimensional fibers. 

\subsection{Simulated data -- stacked images}

\begin{figure}
    \centering
    \includegraphics[width = \linewidth]{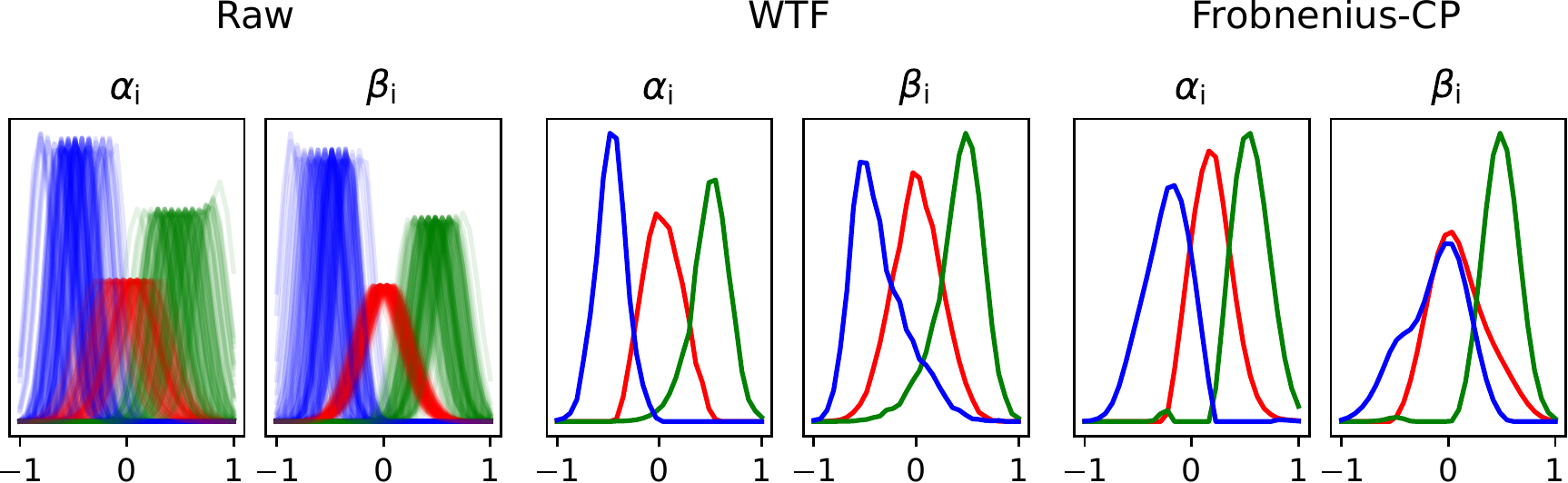}
    \caption{(Left) Marginal distributions $\{\alpha^{(i)}_k, \beta^{(i)}_k : k = 1, 2, 3\}_{i = 1}^{100}$, where $k$ is indexed by color, after applying random translational noise. (Center and right) Atoms learned by WTF and Frobenius-CP decompositions, respectively, as univariate distributions that generate the rank-1 bivariate atoms.  }
    \label{fig:3mode_raw}
\end{figure}

We construct a tensor $X$ of dimensions $100 \times 32 \times 32$, for which the $i$th slice $X_{i, \cdot, \cdot}$ is a $32 \times 32$ discrete distribution constructed as a mixture of three separable bivariate distributions $X_{i, \cdot, \cdot} = {Z_i}^{-1} \sum_{k = 1}^3 \alpha_k^{(i)} \otimes \beta_k^{(i)}$,
where $\{ \alpha^{(i)}_k, \beta^{(i)}_k \}_{k = 1}^3$ are discretised Gaussians on \texttt{linspace(-1, 1, 32)}, and $Z_i$ is a normalising constant. The observed univariate distributions $\{ \alpha^{(i)}_k, \beta^{(i)}_k \}_{k = 1}^3$ are constructed by applying random, normally-distributed translations to some fixed `ground truth' distributions $\{ \alpha_k, \beta_k \}_{k = 1}^3$. We illustrate in Figure \ref{fig:3mode_raw} the simulated dataset, showing the randomly shifted marginal distributions that generate the observed bivariate distributions. Examining the averaged distribution reveals that the noisy observations fluctuate around three modes located along the diagonal, corresponding to the `ground truth' distributions $\alpha_1 \otimes \beta_1, \alpha_2 \otimes \beta_2$, and $\alpha_3 \otimes \beta_3$. 

To find a rank-3 CP decomposition, we applied a standard non-negative CP factorisation with a Frobenius loss as well as WTF with $\varepsilon = 0.01, \rho_i = 0.01, \lambda = 25$. Both methods learn a decomposition where each slice $X_{i, \cdot, \cdot}$ is represented as a mixture of $r = 3$ rank-1 matrices. For WTF, we imposed the additional constraint that the learned univariate atoms lie in the simplex. We show the separable atoms learned by the respective methods also in Figure \ref{fig:3mode_raw}. For ease of comparison, we show normalised atoms in the case of Frobenius-CP.

It is clear that the atoms learned by WTF provide a reasonable summary of the input data. The presence of three distinct unimodal atoms in each dimension is evident, and the spatial arrangement agrees with the distribution of the noisy inputs. On the other hand, Frobenius-CP appears to struggle with the presence of translational noise due to the pointwise nature of the loss function, yielding two atoms in the second dimension that share a mode. 

\begin{figure}
    \includegraphics[width = 0.49\linewidth]{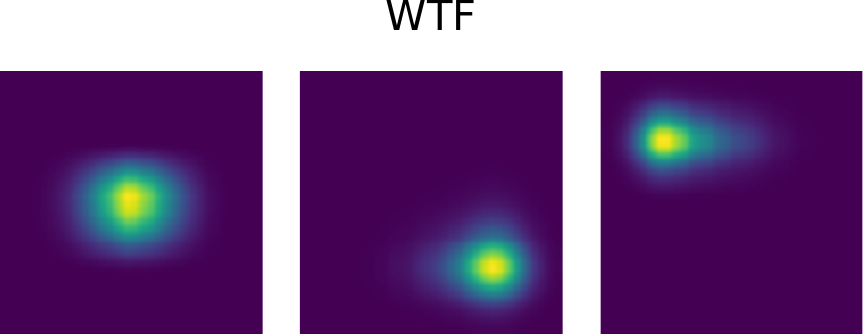}
    \centering
    \includegraphics[width = 0.49\linewidth]{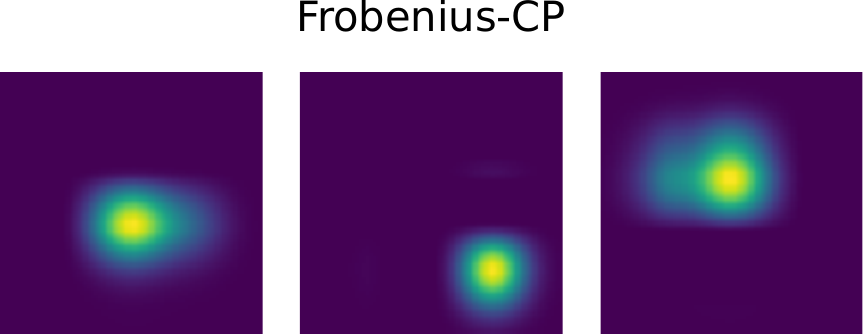} \\ 
    \includegraphics[width = 0.49\linewidth]{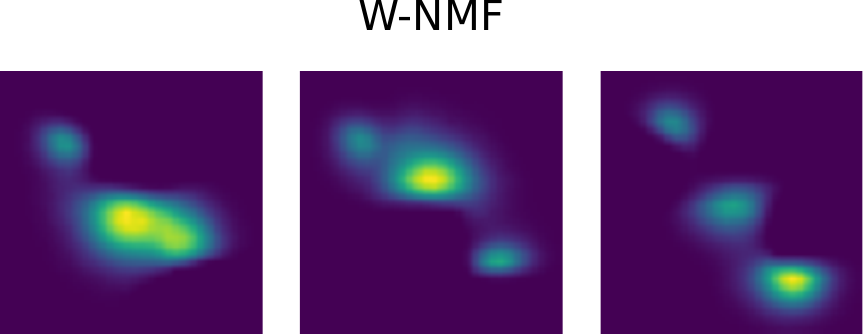}
    \includegraphics[width = 0.49\linewidth]{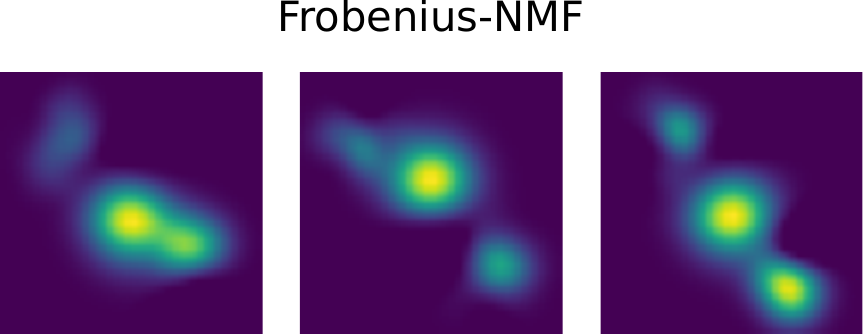}
    \caption{Atoms learned by WTF, Frobenius-CP, W-NMF and Frobenius-NMF.}
    \label{fig:3mode_atoms_2d}
\end{figure}

As an alternative to seeking tensor decompositions, we could vectorise the $32 \times 32$ matrices as columns to form a $1024 \times 100$ matrix and then apply NMF with Frobenius and Wasserstein losses (Frobenius-NMF and WNMF) respectively. For WNMF we used the same parameters as for WTF previously, and require that the components be normalised. We show in Figure \ref{fig:3mode_atoms_2d} the atoms learned by WTF, Frobenius-CP, WNMF and Frobenius-NMF respectively. From this it is clear that the atoms found by matrix factorisation are high-rank, each capturing partial information across all three components of the mixture. In contrast, the atoms found by tensor factorisation are separable, and each atom clearly corresponds to only a single mode.

\subsection{Learning basis for faces}

The AT\&T Olivetti faces dataset \footnote{this dataset is accessible at \url{http://www.cs.nyu.edu/~roweis/}} consists of 400 images (40 subjects, 10 images per subject). Images were resized to $32 \times 32$ and normalised to have unit mass. The dataset was randomly split into a set of training and test images, each of which contained $200$ images (5 images per individual in each set). We constructed from this $200 \times 32 \times 32$ tensors $X_\mathrm{train}$ and $X_\mathrm{test}$ by stacking images from the respective sets as slices along the first mode. WTF was applied to the training data $X_\mathrm{train}$ to find CP decompositions of varying rank $r \in \{10, 20, \ldots, 100 \}$ with parameters $\varepsilon = 10^{-3}, \rho_i = 5 \times 10^{-3} \times r^{-1}, \lambda = 10$, with the constraint that learned atoms lie in the simplex. We also applied a standard non-negative CP decomposition with a squared Frobenius norm loss (which we denote F-CP).

\begin{figure}
    \centering
    \subfloat[]{\includegraphics[width = 0.475\linewidth]{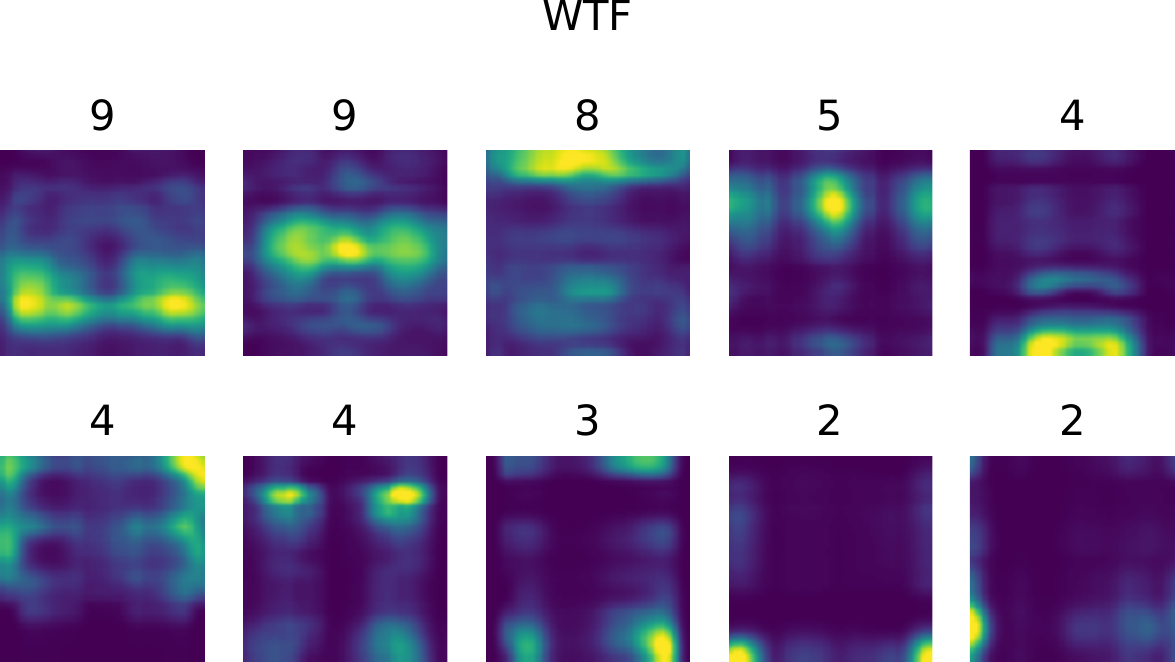}} \hspace{0.5em}
    \subfloat[]{\includegraphics[width = 0.475\linewidth]{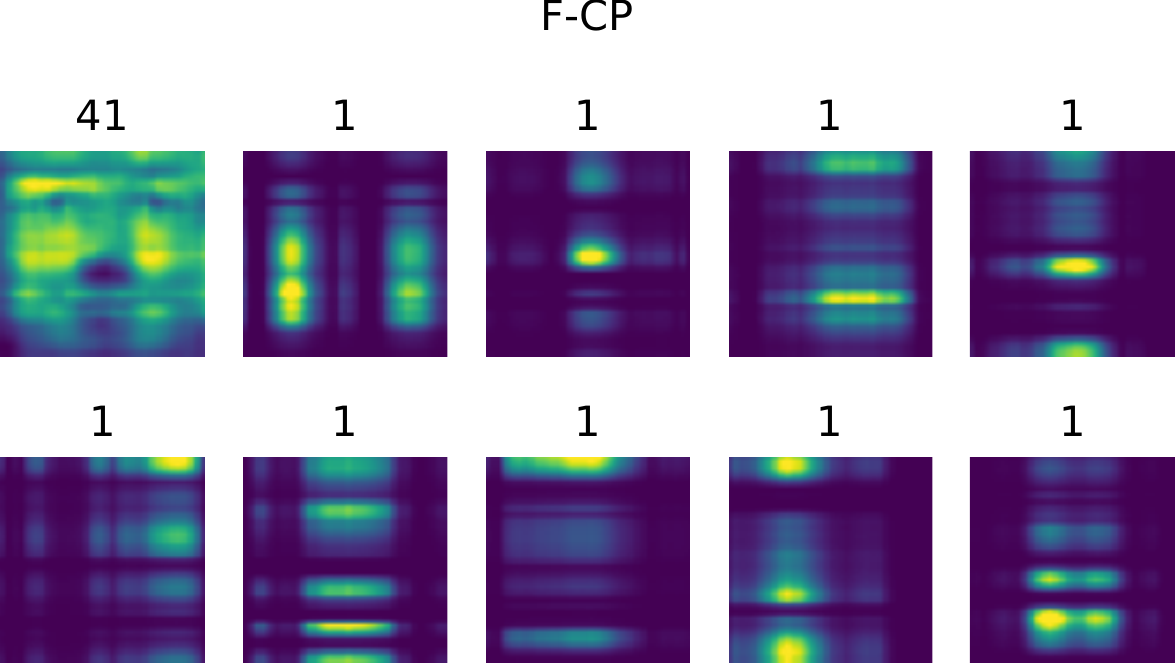}}  \\ 
    \subfloat[]{\includegraphics[width = 0.475\linewidth]{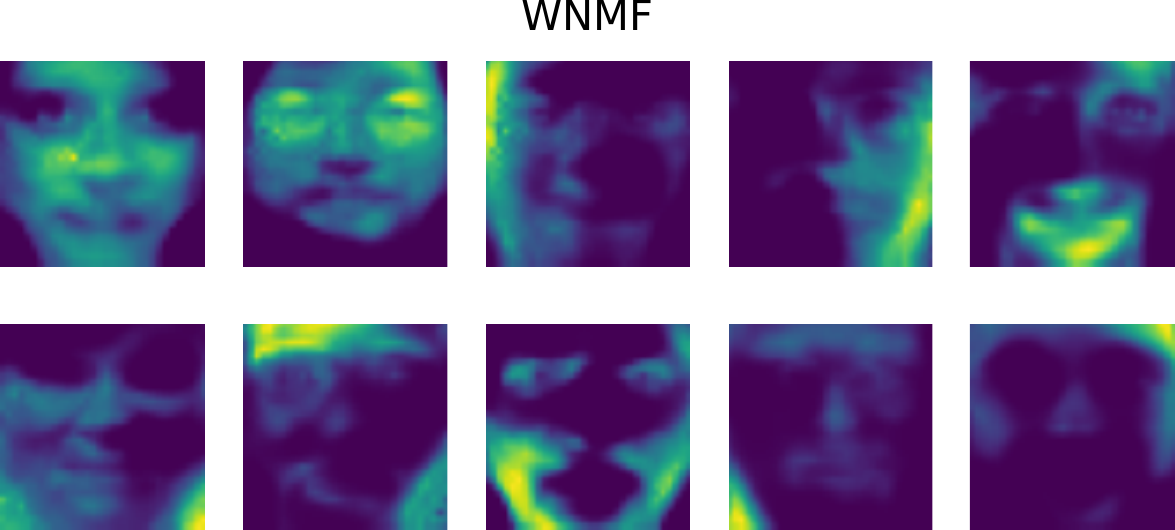}} \hspace{0.5em}
    \subfloat[]{\includegraphics[width = 0.475\linewidth]{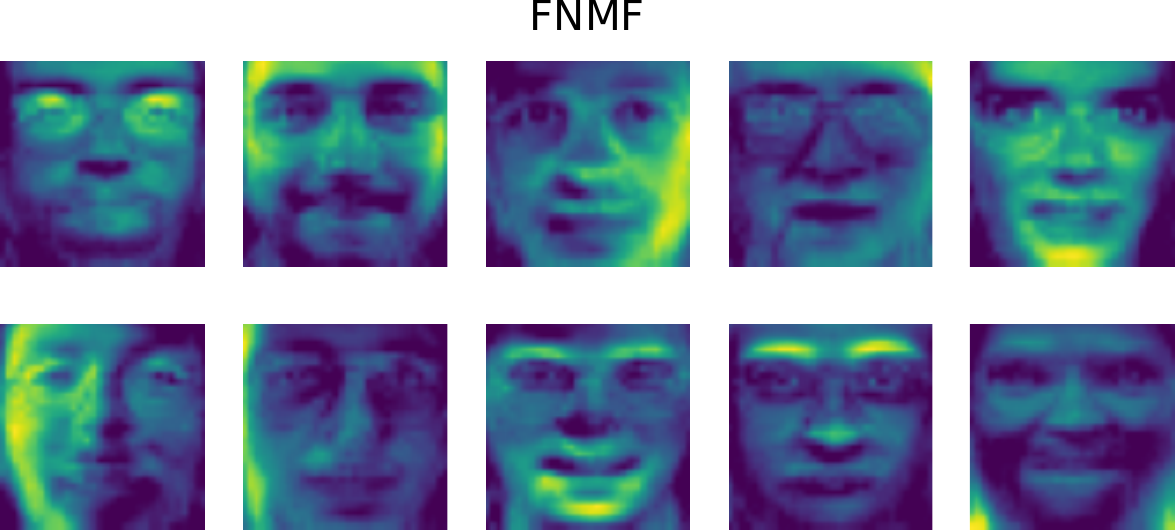}} 
    \caption{(a, b) Superpositions of rank-1 atoms after clustering for finding a tensor approximation of rank 50. The numbers display the number of rank-1 atoms in each cluster. (c, d) Full-rank atoms found by WNMF and Frobenius-NMF respectively, seeking a basis of size 10.}
    \label{fig:faces_clust_atoms}
\end{figure}

To examine the learned factors, motivated by the observations of \cite{hazan2005sparse, shashua2001linear} we expect the separable basis elements to roughly correspond to spatially localised features of the input. To investigate this, we applied spectral clustering to the basis images learned by WTF and Frobenius-CP respectively, and show their superpositions by cluster in Figure \ref{fig:faces_clust_atoms}(a-b). We observe that atoms learned by WTF can be grouped into clusters that highlight spatial regions corresponding to prominent features of the face, such as forehead, cheekbone, nose, etc. On the other hand, the atoms learned by Frobenius-CP effectively fail to cluster, suggesting that each the atoms do not spatially segregate into distinct features.

\begin{figure}
    \centering
    \subfloat[Tensor factorisations]{\includegraphics[width = 0.49\linewidth]{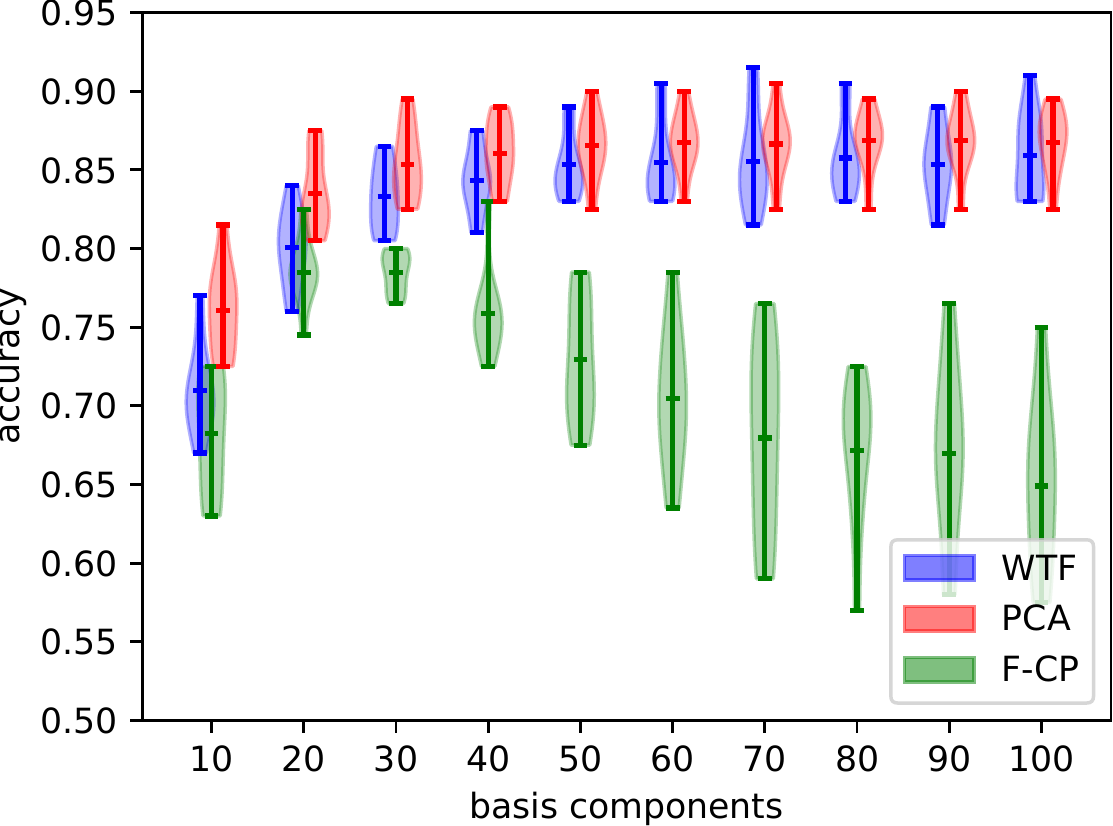}}
    \subfloat[Matrix factorisations]{\includegraphics[width = 0.49\linewidth]{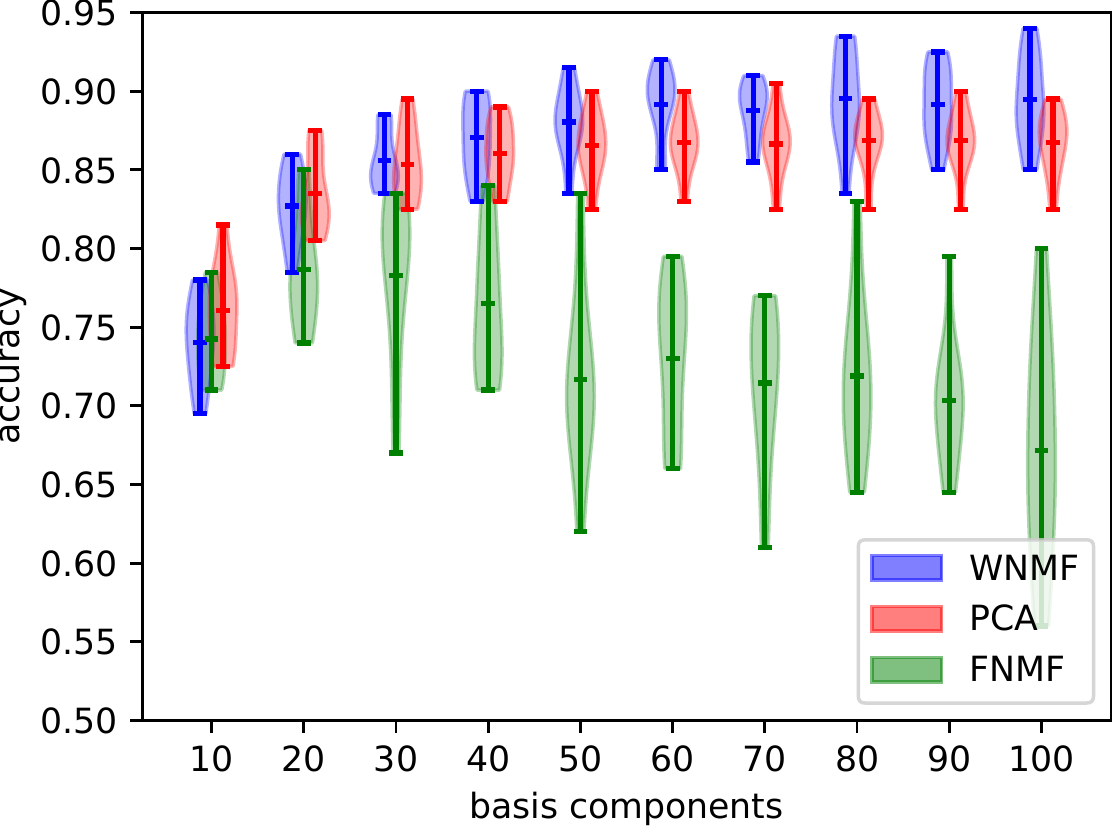}}
    \caption{Classification accuracy as a function of the number of basis components for (a) tensor factorisations (separable atoms) and (b) matrix factorisations (full rank atoms).}
    \label{fig:faces_acc}
\end{figure}

To assess the usefulness of the learned basis for supervised classification, we projected the test dataset $X_\mathrm{test}$ onto the basis learned from the training set. This was done by solving \eqref{eq:dual_factors} for a coefficient matrix $A^{(1)}_\mathrm{test}$ whilst holding the factor matrices encoding atoms $\{ A^{(2)}, A^{(3)} \}$ fixed. This problem is convex, so we are guaranteed a unique solution. The rows of $A^{(1)}_\mathrm{test}$ are the coordinates of the images in the basis $\{ A^{(2)}, A^{(3)} \}$ learned from the training dataset. Following \cite{sandler2011nonnegative}, we then use 1-nearest neighbour classification with a cosine distance $(x, y) \mapsto 1 - \cos(\angle(x, y))$ to assign each test image to one of the 40 individual labels. In Figure \ref{fig:faces_acc}(a) we summarise the accuracy of this approach over 10 random train-test splits as a function of the basis size (factorisation rank) $r$. As a reference for performance, we also display the accuracy for a basis of $r$ principal components (PCs). Note that the PCs are full-rank $32 \times 32$ matrices that are not non-negative, compared to the rank-1 non-negative basis atoms sought by tensor factorisation methods. Thus, each full-rank atom contains 16-fold more entries than a rank-1 atom. We find that classification using the WTF basis learned from $X_\mathrm{train}$ achieves performance comparable to the PCA basis, despite the additional restrictions for WTF that the basis elements be rank-1 and non-negative. In contrast, the performance of Frobenius-CP degrades as the number of components increases, suggesting that the basis elements found using the pointwise Frobenius loss have a poor ability to generalise beyond the training examples. 

We next compare the tensor representation found by WTF against the equivalent representation found by matrix decompositions using Wasserstein-NMF (W-NMF). Training and test datasets were constructed as in the tensor case, except we took $X_\mathrm{train}$ and $X_\mathrm{test}$ to be $1024 \times 200$ matrices with columns corresponding to the vectorised images. We sought a decomposition using both Frobenius (F-NMF) and Wasserstein (W-NMF) losses. For W-NMF, we used parameters identical to WTF. Each atom is a vector of length $1024$ which represents a $32 \times 32$ matrix with no constraints on rank, in contrast to the rank-1 constraint in the case of the tensor representation. 

In Figure \ref{fig:faces_clust_atoms}(c-d) we show the individual atoms found by WNMF and Frobenius-NMF respectively. Curiously, as in the case of tensor factorisations, the atoms found by WNMF visually correspond to localised facial features. In contrast, all of the atoms found by FNMF redundantly capture the full structure of the face. We assessed the performance of the bases found by WNMF and FNMF for classification of the test dataset $X_\mathrm{test}$. As shown in Figure \ref{fig:faces_acc}(b), we find that WNMF achieves a classification accuracy that is on-par or higher than PCA. On the other hand, as in the case of tensor decompositions, the accuracy of FNMF decreases as the number of basis components is increased. Finally, we note that for a fixed number of basis components $r$, the matrix representation requires effectively 4.6 fold more stored entries than the CP tensor representation, and each full-rank atom is equivalent to 16 rank-1 entries in terms of stored entries. However, for the same number of basis elements we find that WTF achieves a classification accuracy that is comparable to WNMF. This suggests that the tensor format is more efficient for representing image data \cite{hazan2005sparse, shashua2001linear}. 

\section{Conclusion}

Motivated by practical settings where observed data lie on a space with metric structure, we formulated the problem of finding non-negative factorisations of matrices and tensors using a Wasserstein loss and propose to solve it numerically via the dual formulation. 
Alone the way, we derived a closed-form Legendre transform for the semi-unbalanced Wasserstein loss \eqref{eq:semiunbal_ot_def} 
, which to our knowledge has not been previously reported in the literature. 
Avenues for future work include generalising our approach to deal with sparse data as in \cite{afshar2020swift}, as well as exploring alternative choices of barrier functions for the non-negativity constraint. One direction of interest is to develop a methodology where the operation of taking linear combinations of atoms is replaced with taking the Wasserstein barycenter \cite{cuturi2014fast}, as was done in the setting of matrix factorisations by Schmitz et al. \cite{schmitz2018wasserstein}. 

\section*{Acknowledgements}

S.Z. would like to thank Elina Robeva for an introduction to the theory of tensors, Hugo Lavenant for an introduction to the duality theory of smoothed optimal transport, and Igor Pinheiro for insightful discussions. 

{\small
\bibliographystyle{ieee_fullname}
\bibliography{references}
}

\pagebreak
\onecolumn
\setcounter{section}{0}
\renewcommand\thesection{\Alph{section}}

\section{Introduction of notation}\label{sec:notation}

We summarise below the (relatively standard) notation conventions which we adopt throughout this article. 
\begin{itemize}
    \item Matrices and tensors are denoted in upper case, e.g. $X, U, V$, in contrast to vectors which are written in lower case, e.g. $x, u , v$. 
    \item For a matrix $X \in \mathbb{R}^{m \times n}$, we index its elements $X_{ij}$ and write $X_i$ for the $i$th column as a $m$-dimensional vector. 
    \item For a tensor $X \in \mathbb{R}^{n_1 \times \cdots \times n_d}$, we index its elements $X_{i_1, \ldots, i_d}$ and write $X_{(i)}$ for its matricisation \cite{kolda2009tensor} along mode $i$, which is a matrix of dimensions $n_i \times n_1 \cdots n_{i-1} n_{i+1} \cdots n_d$. 
    \item We write the inner product for vectors $(x, y)$ as $\inner{x, y} = \sum_i x_i y_i$, for matrices $(X, Y)$ as $\inner{X, Y} = \sum_{ij} X_{ij} Y_{ij}$ and so on for tensors. 
    \item We denote elementwise multiplication by $\odot$, and outer product of vectors by $\otimes$. Unless otherwise specified, for $x, y$ vectors and $A$ a matrix of appropriate dimensions, by writing $Ax$ we refer to the matrix-vector product, and by $xy$ and $x/y$ we refer to the elementwise product and quotient respectively.
    \item Following the notation of \cite{kolda2009tensor}, we write the mode-$k$ product of a tensor $A \in \bbR^{n_1 \times \cdots \times n_d}$ and a matrix $B \in \bbR^{m \times n_k}$ as 
    \begin{align*}
        &(A \times_k B)_{i_1, \ldots, i_{k-1}, j, i_{k+1}, \ldots, i_d} \\
            &\hspace{1.5em} = \sum_{i_k = 1}^{n_k} A_{i_1, \ldots, i_d} B_{j, i_k},
    \end{align*}
    and this is a tensor of dimensions $n_1 \times \cdots \times n_{k-1} \times m \times n_{k+1} \times \cdots \times n_d$. 
    \item For non-negative matrices or tensors $\alpha$ and $\beta$, we write the entropy as $E(\alpha) = \inner{\alpha, \log(\alpha) - 1}$, the relative entropy $\Ent(\alpha | \beta) = \inner{\alpha, \log(\alpha/\beta) - 1}$ and the generalised Kullback-Leibler divergence as $\KL(\alpha | \beta) = \inner{\alpha, \log(\alpha/\beta)} - \inner{\alpha, \ones} + \inner{\beta, \ones}$. 
    \item For a discrete metric space $\mathcal{X}$, we write $\mathcal{P}(\mathcal{X})$ and $\mathcal{M}_+(\mathcal{X})$ to be respectively the set of probability distributions and positive measures supported on $\mathcal{X}$.
\end{itemize}

\section{Wasserstein-NMF as a special case of WTF}\label{sec:wnmf}
From the framework introduced in Sections \ref{sec:factors} and \ref{sec:core}, we may recover the NMF method described by Rolet et al. \cite{rolet2016fast} when we consider matrices as 2-way tensors. Let $X \in \mathbb{R}^{m \times n}$. In the context of NMF, columns of $X$ correspond to observations, so we may take $\Phi(X, \hat{X})$ to be a Wasserstein loss along the columns of its arguments. 
We take $S_{i_1, i_2} = \delta_{i_1, i_2}$ to be fixed, and so for two factor matrices $U, V$ we have 
\begin{align*}
    S[U, V] &= \sum_{i = 1}^r U_i \otimes V_i = UV^\top. 
\end{align*}
Thus, \eqref{eq:wtf_general_smooth} becomes 
\begin{align*}
    \min_{U, V} \Phi(X, UV^\top) + \rho_1 E_{\Sigma_1}(U) + \rho_2 E_{\Sigma_2}(V),
\end{align*}
which coincides (albeit with differing notation) with the Wasserstein-NMF problem introduced by Rolet et al. \cite{rolet2016fast}. Proposition \ref{prop:factor_duality} gives the dual problem for the subproblems in factor matrices $U$ and $V$ respectively.  

\section{Convex duality}\label{sec:convex_duality}

For further background on the techniques involved, we refer the reader to the existing literature on variational problems involving optimal transport \cite{cuturi2016smoothed, cuturi2018semidual, frogner2015learning}. 
\begin{defn}[Legendre transform]
    Let $f: \mathbb{R}^n \to (-\infty, \infty]$ be a proper function, i.e. one that is not identically $+\infty$. The Legendre transform of $f$ is defined for $u \in \mathbb{R}^n$ as 
    \begin{align*}
        f^*(u) = \sup_{x \in \mathbb{R}^n} \inner{x, u} - f(x).
    \end{align*}
    Furthermore, $f^{**} = f$ if and only if $f$ is convex and lower semicontinuous. 
\end{defn}

\begin{theorem}[Fenchel-Rockafellar theorem {\cite{rockafellar1967}}]\label{thm:fenchel}
    Let $E, F$ be (finite or infinite dimensional) real vector spaces, and $E^*, F^*$ their respective topological dual spaces. Let $f : E \to (-\infty, \infty]$, $g : F \to (-\infty, \infty]$ be proper (not identically $+\infty$), convex, lower-semicontinuous (sublevel sets are closed) functions. Let $A : E \to F$ be a continuous linear operator. Consider the convex minimisation problem 
    \begin{align*}
        \min_{x \in E} f(x) + g(Ax) \tag{P}
    \end{align*}
    Then (P) has a corresponding dual problem (P*)
    \begin{align*}
        \sup_{y \in F^*} -f^*(A^* y) - g^*(-y), \tag{P*}
    \end{align*}
    where $A^*$ is the adjoint of $A$, and $f^*(\cdot) = \sup_{x \in E} \inner{x, \cdot} - f(x)$, $g^*(\cdot) = \sup_{y \in F} \inner{y, \cdot} - g(y)$ are the \emph{Legendre transforms} of $f$ and $g$, defined over $E^*$ and $F^*$ respectively. 
\end{theorem}

In general the dual problem (P*) provides a lower bound on the solution to the primal problem (P): $p^\star \ge d^\star$. However, the following simple condition is sufficient for equality to hold in a finite dimensional setting.

\begin{theorem}[Condition for strong duality in finite dimensions, adapted from {\cite{rockafellar1967}}]
    If $E, F$ are finite-dimensional and there exists some $x$ in the relative interior of the feasible set, then $p^\star = d^\star$.
\end{theorem}

\section{Proofs}\label{sec:proofs}

\begin{proof}[Proof of Proposition \ref{prop:factor_duality}]
    We write the primal problem as \eqref{eq:primal_factors}:
    \begin{align*}
        \min_{A^{(k)}} \Phi(X, S[A^{(1)}, \ldots, A^{(d)}]) + \rho_k E_{\Sigma_k}(A^{(k)}).
    \end{align*}
    Now let us substitute the definition of the Legendre transform of $\Phi$, and we take formally an $\inf - \sup$ exchange:
    \begin{align*}
        &\min_{A^{(k)}} \sup_U \left[ \inner{U, S \times_1 A^{(1)} \times \cdots \times_d A^{(d)}} - \Phi^*(X, U) \right] + \rho_k E_{\Sigma_k}(A^{(k)}) \\
        &= \sup_U -\Phi^*(X, U) + \min_{A^{(k)}} \inner{U, S \times_1 A^{(1)} \times \cdots \times_d A^{(d)}} + \rho_k E_{\Sigma_k}(A^{(k)}) \\
        &= \sup_U -\Phi^*(X, U) - \rho_k \max_{A^{(k)}} \left[ \frac{-1}{\rho_k}\inner{U,  S \times_1 A^{(1)} \times \cdots \times_d A^{(d)}} - E_{\Sigma_k}(A^{(k)}) \right].
    \end{align*}
    We now note the identities $\inner{A, B \times_k C} = \inner{A \times_k C^\top, B}$ and $\inner{A \times_k B, C} = \inner{BA_{(k)}, C_{(k)}}$ so:
    \begin{align*}
        \inner{U,  S \times_1 A^{(1)} \times \cdots \times_d A^{(d)}} &= \inner{U \times_{j \ge k+1} (A^{(j)})^\top, S \times_{j \leq k} A^{(j)}} \\
        &= \inner{U \times_{j \ge k+1} (A^{(j)})^\top, S \times_{j \leq k-1} A^{(j)} \times_k A^{(k)}} \\
        &= \inner{A^{(k)} \left[ S \times_{j \leq k-1} A^{(j)} \right]_{(k)} , \left[ U \times_{j \ge k+1} (A^{(j)})^\top \right]_{(k)}} \\
        &= \inner{A^{(k)} , \left[ U \times_{j \ge k+1} (A^{(j)})^\top \right]_{(k)}  \left[ S \times_{j \leq k-1} A^{(j)} \right]_{(k)}^\top } \\
        &= \inner{A^{(k)}, \Xi^{(k)}(U)},
    \end{align*}
    where $\Xi^{(k)}(U) = \left[ U \times_{j \ge k+1} (A^{(j)})^\top \right]_{(k)}  \left[ S \times_{j \leq k-1} A^{(j)} \right]_{(k)}^\top$. Thus,
    \begin{align*}
        &\max_{A^{(k)}} \frac{-1}{\rho_k}\inner{U,  S \times_1 A^{(1)} \times \cdots \times_d A^{(d)}} - E_{\Sigma_k}(A^{(k)}) \\
        &= \max_{A^{(k)}} \inner{A^{(k)}, \frac{-1}{\rho_k} \Xi^{(k)}(U)} - E_{\Sigma_k}(A^{(k)}) \\  
        &= E_{\Sigma_k}^* \left( \frac{-1}{\rho_k} \Xi^{(k)}(U) \right).
    \end{align*}
    Thus we have 
    \begin{align*}
        \sup_U -\Phi^*(X, U) - \rho_k E_{\Sigma_k}^* \left( \frac{-1}{\rho_k} \Xi^{(k)}(U) \right).
    \end{align*}
    Strong duality holds by application of the Fenchel-Rockafellar theorem. 
    
    Let the value of $U$ at optimality be $U^\star$. Then the corresponding factor matrix $(A^{(k)})^\star$ must be the solution of 
    \begin{align*}
        \max_{A^{(k)}} \inner{A^{(k)}, \frac{-1}{\rho_k} \Xi^{(k)}(U^\star)} - E_{\Sigma_k}(A^{(k)}).
    \end{align*}
    If $\Sigma_k = \{ \}$, then $E_{\Sigma_k}(x) = \inner{x, \log(x) - 1}$. We are unconstrained and have the sum of an affine and a convex term. Differentiating, we find the first-order optimality condition 
    \begin{align*}
        {A^{(k)}}^\star = \exp\left( \frac{-1}{\rho_k} \Xi^{(k)}(U^\star) \right).
    \end{align*}
    If $\Sigma_k = \{ A^{(k)} : \inner{A^{(k)}, \ones} = 1 \}$, then the problem is subject to a simplex constraint. At optimality, therefore, the gradient must be orthogonal to the simplex (parallel to $\ones$):
    \begin{align*}
        \frac{-1}{\rho_k} \Xi^{(k)}(U^\star) - \log({A^{(k)}}^\star) &= c\ones \\
        \implies {A^{(k)}}^\star &= \exp(-c) \exp\left( \frac{-1}{\rho_k} \Xi^{(k)}(U^\star) \right).
    \end{align*}
    Since $\inner{A^{(k)}, \ones} = 1$, we conclude that $\exp(c) = \exp\left( \frac{-1}{\rho_k} \Xi^{(k)}(U^\star) \right)$. 
    In the cases where $\Sigma_k$ requires row or column normalisation, applying an identical argument row- or columnwise leads to an analogous result, where we normalise the output row- or columnwise. 
\end{proof}

\begin{proof}[Proof of Proposition \ref{prop:core_duality}]
    As in the proof of Proposition \ref{prop:factor_duality}, we introduce the Legendre transform of $\Phi$ and carry out an $\inf-\sup$ exchange.
    \begin{align*}
        &\min_S \Phi(X, S[A^{(1)}, \ldots, A^{(d)}]) + \rho_0 E_{\Sigma_0}(S) \\
        &= \min_S \sup_U \left[ \inner{U, S \times_1 A^{(1)} \times \cdots \times_d A^{(d)}} - \Phi^*(X, U) \right] + \rho_0 E_{\Sigma_0}(S) \\
        &= \sup_U -\Phi^*(X, U) + \min_S \left[ \inner{U, S \times_1 A^{(1)} \times \cdots \times_d A^{(d)}} + \rho_0 E_{\Sigma_0}(S) \right] \\
        &= \sup_U -\Phi^*(X, U) - \rho_0 \max_S \left[ \frac{-1}{\rho_0} \inner{U, S \times_1 A^{(1)} \times \cdots \times_d A^{(d)}} - E_{\Sigma_0}(S) \right].
    \end{align*}
    Now note that using the identities presented in the proof of Proposition \ref{prop:factor_duality}:
    \begin{align*}
        \inner{U, S \times_1 A^{(1)} \times \cdots \times_d A^{(d)}} &= \inner{U \times_1 (A^{(1)})^\top \times \cdots \times_d (A^{(d)})^\top, S} \\
        &= \inner{S, \Omega(U)}
    \end{align*}
    where $\Omega(U) = U \times_1 (A^{(1)})^\top \times \cdots \times_d (A^{(d)})^\top$. Thus,
    \begin{align*}
        &\max_S \frac{-1}{\rho_0} \inner{U, S \times_1 A^{(1)} \times \cdots \times_d A^{(d)}} - E_{\Sigma_0}(S) \\
        &= \max_S \inner{S, \frac{-1}{\rho_0}\Omega(U)} - E_{\Sigma_0}(S) \\
        &= E_{\Sigma_0}^* \left( \frac{-1}{\rho_0}\Omega(U) \right).
    \end{align*}
    Thus, the dual problem is 
    \begin{align*}
        \sup_U -\Phi^*(X, U) - \rho_0 E_{\Sigma_0}^* \left( \frac{-1}{\rho_0}\Omega(U) \right). 
    \end{align*}
    As before, strong duality holds by application of the Fenchel-Rockafellar theorem. 
    
    Let the value of $U$ at optimality be $U^\star$. Then the corresponding core tensor $S^\star$ must be the solution of 
    \begin{align*}
        \max_S \inner{S, \frac{-1}{\rho_0}\Omega(U^\star)} - E_{\Sigma_0}(S).
    \end{align*}
    Following the previous argument given in the Proof of Proposition \ref{prop:factor_duality}, we find that 
    \begin{align*}
        S^\star &= \begin{cases} 
            \exp\left( \frac{-1}{\rho_0} \Omega(U^\star) \right), &\Sigma_0 = \{ \}, \\ 
            \frac{\exp\left( \frac{-1}{\rho_0} \Omega(U^\star) \right)}{\inner{\exp\left( \frac{-1}{\rho_0} \Omega(U^\star) \right), \ones}}, &\Sigma_0 = \{ S : \inner{S, \ones} = 1 \}
        \end{cases}.
    \end{align*}
\end{proof}

\begin{proof}[Proof of Proposition \ref{prop:semiunbal_dual}]
For marginal distributions $p, q$, we write an alternative form of the semi-balanced optimal transport problem \eqref{eq:semiunbal_ot_def} as 
\begin{align*}
    \OT_{\varepsilon}^{\lambda}(p, q) &= \inf_{\gamma : \gamma \ones = p} \varepsilon \Ent(\gamma | K) + \lambda \KL(\gamma^\top \ones | q).
\end{align*}
We seek the Legendre transform in the second argument $q$. Introduce $u$ the dual variable of $q$ and $\alpha$ the Lagrange multiplier for the constraint $\gamma \ones = p$, then exchange the inf and sup. 
\begin{align*}
    {\OT_{\varepsilon}^{\lambda}}^*(p, u) &= \sup_{q} \inner{u, q} - \inf_{\gamma \ones = p} \left[ \varepsilon \Ent(\gamma | K) + \lambda \KL(\gamma^\top \ones | q) \right] \\
            &= \inf_\alpha \sup_{q, \gamma} \inner{u, q} - \varepsilon \Ent(\gamma | K) \\
            &\hspace{5em} - \lambda \KL(\gamma^\top \ones | q) + \inner{\alpha, \gamma \ones - p}.
\end{align*}
Use first order condition for $q$:
\begin{align*}
    \frac{\partial}{\partial q}(\cdot) &= u - \lambda \left( 1 - \frac{\gamma^\top \ones}{q} \right) = 0 \\
    \Rightarrow q &= (\gamma^\top \ones) \left( \frac{\lambda}{\lambda - u} \right),
\end{align*}
where multiplication is elementwise. Substituting back, we find that 
\begin{align*}
    \begin{split}
        \inf_{\alpha} &\sup_{\gamma} \inner{u, \left( \gamma^\top \ones \odot \frac{\lambda}{\lambda-u} \right)} - \varepsilon \Ent(\gamma | K) \\
        &- \lambda \inner{\gamma^\top \ones, \log\left( \frac{\lambda - u}{\lambda} \right) - \gamma^\top \ones + q } + \inner{\alpha, \gamma \ones - p}.
    \end{split}
\end{align*}
Differentiating with respect to $\gamma$, (and after some involved algebra) we find the first order condition for $\gamma$ to be 
\begin{align*}
    \gamma_{ij} &= \exp\left( \frac{\alpha_i}{\varepsilon} \right) K_{ij} \left( \frac{\lambda}{\lambda - u_j} \right)^{\lambda/\varepsilon}. 
\end{align*}
Substituting back and finally differentiating in $\alpha$ we find that 
\begin{align*}
    \alpha &= \varepsilon \log\left( \frac{p}{K(\frac{\lambda}{\lambda - u})^{\lambda/\varepsilon}} \right) = \varepsilon \log\left( \frac{p}{Kf} \right),
\end{align*}
where $f = \left( \frac{\lambda}{\lambda - u} \right)^{\lambda/\varepsilon}$ for brevity. With all this, the Legendre transform is 
\begin{align*}
    {\OT_{\varepsilon}^{\lambda}}^*(p, u) &= -\inner{p, \varepsilon \log\left(\frac{p}{Kf}\right)} + \inner{K^\top \frac{p}{Kf}, \varepsilon f} \\
    &= -\varepsilon\inner{p, \log\left( \frac{p}{Kf} \right)} + \varepsilon \inner{p, \ones}.
\end{align*}
The relationship between the primal and dual variables is therefore
\begin{align*}
    q &= \left( \frac{\lambda}{\lambda - u} \right) f \odot K^\top \frac{p}{Kf}, \\
    \gamma &= \exp\left(\frac{\alpha}{\varepsilon}\right) K f.
\end{align*}
\end{proof}

\section{Gibbs kernel convolution}\label{sec:conv}

\begin{prop}[Convolution with tensor-valued Gibbs kernel]
    In the context of Proposition \ref{prop:loss_tensor}, the Gibbs kernel $K = e^{-C/\varepsilon}$ is a $2d$-way tensor that decomposes multiplicatively: 
    \begin{align*}
        K_{i_1, \ldots, i_d, j_1, \ldots, j_d} &= e^{-C_{i_1, \ldots, i_d, j_1, \ldots, j_d}/\varepsilon} \\
        &= K^{(1)}_{i_1, j_1} \cdots K^{(d)}_{i_d, j_d}.
    \end{align*}
    Furthermore, note that in the case of vector-valued input, the formula \eqref{eq:semiunbal_ot_legendre} involves a matrix-vector convolution of the form $s \mapsto Ks$. In our setting, for $S \in \mathbb{R}^{n_1 \times \cdots \times n_d}$ the corresponding operation is a convolution along all modes that has the following decomposition:
    \begin{align*}
        (KS)_{i_1, \ldots, i_d} &= \sum_{j_1, \ldots, j_d} K_{i_1, \ldots, i_d, j_1, \ldots, j_d} S_{j_1, \ldots, j_d} \\
        &= S \times_{i = 1}^d K^{(i)}.
    \end{align*}
\end{prop}

\section*{Code}

An implementation of the methods described in this paper is available at \url{https://github.com/zsteve/wtf}

\end{document}